%% file: ms.tex
\documentclass[10pt,twocolumn,letterpaper]{article}

\usepackage[final]{pdfpages}
\usepackage{cvpr}
\usepackage{times}
\usepackage{epsfig}
\usepackage{graphicx}
\usepackage{amsmath}
\usepackage{amssymb}
\usepackage{enumitem}
\usepackage{blindtext}
\usepackage{subfigure}
\usepackage[table,x11names,dvipsnames,table]{xcolor}

\definecolor{myorchid}{RGB}{150,10,30}
\definecolor{myblue}{RGB}{10,30,250}
\definecolor{mygreen}{RGB}{10,120,10}

 \newcommand{\ADDED}[1]{\color{black}{#1}}
 \newcommand{\REMOVED}[1]{}
\newcommand{\asmr}{YouTube-ASMR}
\renewcommand{\paragraph}[1]{\vspace{0.03cm} \noindent{\bf #1}}

\usepackage[pagebackref=true,breaklinks=true,letterpaper=true,colorlinks,bookmarks=false]{hyperref}

 \cvprfinalcopy 

\def\cvprPaperID{7612} 

\ifcvprfinal\fi
\begin{document}

\title{Telling Left from Right: Learning Spatial Correspondence of Sight and Sound}

\author{Karren Yang\thanks{Work done at Adobe Research during KY’s summer internship.} \\
MIT\\
\and
Bryan Russell \\
Adobe Research\\
{\small \url{http://karreny.github.io/telling-left-from-right}}
\and
Justin Salamon \\
Adobe Research\\
}

\maketitle

\begin{abstract} 
Self-supervised audio-visual learning aims to capture useful representations of video by leveraging correspondences between visual and audio inputs. Existing approaches have focused primarily on matching semantic information between the sensory streams. We propose a novel self-supervised task to leverage an orthogonal principle: matching spatial information in the audio stream to the positions of sound sources in the visual stream. Our approach is simple yet effective. We train a model to determine whether the left and right audio channels have been flipped, forcing it to reason about spatial localization across the visual and audio streams. To train and evaluate our method, we introduce a large-scale video dataset\ADDED{, YouTube-ASMR-300K,} with spatial audio comprising over 900 hours of footage. We demonstrate that understanding spatial correspondence enables models to perform better on three audio-visual tasks, achieving quantitative gains over supervised and self-supervised baselines that do not leverage spatial audio cues. \ADDED{We also show how to extend our self-supervised approach to 360 degree videos with ambisonic audio.}

\end{abstract}

\input{sections/introduction.tex}

\input{sections/related.tex}

\input{sections/stereo.tex}

\input{sections/reduced_ambisonic.tex}

\input{sections/conclusion.tex}

\subsection*{Acknowledgements} The authors would like to thank Andrew Owens for helpful discussions in the early stages of the project.

{\small
\bibliographystyle{ieee_fullname}
\bibliography{egbib}
}

\mbox{~}
\clearpage
\newpage



\section*{Supplementary material}

# Appendices

The Supplemental includes video examples corresponding to Figure 1 of the main text as well as additional details of dataset collection and experiments. Section A describes our collection and curation of the YouTube-ASMR-300K and YouTube-ASMR datasets. Sections B, C and D provide additional details of implementation and baselines for downstream tasks. Finally, Section E describes the implementation details and results of extending the audio-visual spatial correspondence task to 360-degree videos.

## A. YouTube-ASMR Dataset

**Collecting YouTube URLs.** We obtained an initial set of approximately 2K URLs of YouTube videos related to ASMR using the search keywords such as "ASMR Binaural", "ASMR Ear-to-ear", "ASMR Tingles" and "ASMR Tapping". Subsequently, we expanded this list by scraping the URLs of related videos in multiple iterations, obtaining a final list of approximately 80K unique URLs. The average length of each video in this list was approximately 30 minutes long based on metadata information. For each URL, we downloaded up to 10-15 video clips of 10 s duration, for a total initial count of approximately 1M video clips.

**Data filtering.** We filtered the downloaded video clips based on separate criteria for the visual and the audio streams. For the audio stream, we filtered for videos with stereo sound that was perceived to be off-center. In particular, we computed the log difference in the magnitude spectrogram between the left and right stereo channels and produced a weighted average of the difference over the frequency domain based on the amplitude, as proposed by [11]. We then filtered for video clips in which at least 60% of these difference values over time bins was significantly different from zero. For the visual stream, we performed face detection [2] and filtered for video clips that contained a face with high confidence. We also removed videos containing still images by computing differences between frames over different time steps. Following these data filtering steps, the resulting dataset, YouTube-ASMR-300K, consists of over 300K video clips from approximately 30K unique YouTube URLs.

**Manual curation of YouTube-ASMR subset.** For additional quality control, we selected 50 top YouTube channels from the YouTube-ASMR-300K dataset and manually assessed one-third of their videos. We then created a subset of the YouTube-ASMR-300K dataset comprising of the videos from 30 YouTube channels with strong audio-visual spatial correspondence based on our manual inspection.

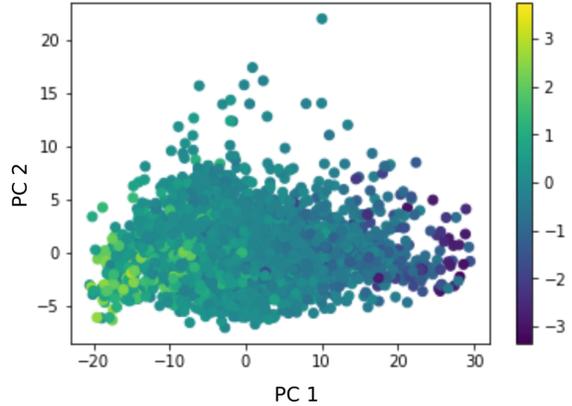

Figure 1. PCA visualization of audio embeddings extracted from YouTube-ASMR dataset, colored by log energy-difference between left and right audio channels. Each point corresponds to a different video in the held-out data.

## B. Sound Localization

**Additional details of correlation analysis.** Here we provide additional information corresponding to the analysis from the beginning of Section 5.1. Recall that we first evaluate whether the audio embeddings learned by our model contain spatial information. We do this by comparing the learned audio features to a sound source's spatial location at each time instance. To obtain an approximate location of a sound source at a time instance, we compute the log-energy difference between the two audio channels. Supplemental Table 1 shows Pearson R and Spearman R correlation coefficients to measure the correlation between the approximate sound location computed from directional energy and projections of the self-supervised learned embedding. We find that the audio features are strongly correlated with the location of the sound source based on both an unsupervised projection (*i.e.*, principal component analysis) and a supervised projection (*i.e.*, canonical correlation analysis) of the learned embeddings. Additionally, Supplemental Figure 1 shows a PCA plot of the audio embeddings colored by the log energy-difference; observe that the first principal component of the embeddings strongly reflects the log energy-difference.

Next, recall that we investigated whether the visual sub-network has learned to identify the positions of sound sources, as a result of having to match them with the spatial cues provided by the audio sub-network. To determine the regions that the visual sub-network attends to, we use a sliding window of 32 pixels and replace regions of the visual frame with their mean values. We then pass the modified frames through our visual sub-network and determine those regions whose omission maximally affects prediction on the pretext task [4, 12]. Supplemental Table 1 shows our

evaluations of the predicted regions of importance using the masking approach of [4, 12] ("Predicted ROI") by comparing them to the approximate ground truth locations based on the log energy-difference between audio channels. We find that the predicted localization results are significantly correlated with the approximate ground truth locations.

**Baselines.** To preclude the possibility that the visual network is relying on artist-dependent spatial biases to achieve this correlation rather than the dynamic spatial audio cues, we generate baseline predictions of sound localization based on the prior distribution of the sound, visual salience, and motion in training videos produced by the same artists using metadata information. All of these baselines exploit biases in the patterns of sound source localization that may be present in different ASMR artists' videos. The observation that our result outperforms these baselines indicates that the sound source localization is being performed based on audio-visual spatial correspondence, rather than these biases.

- **Sound Prior.** For a given test video clip, we consider all training set video clips made by the same YouTube channel. We then take the directional energy of the sound, *i.e.*, the log energy-difference between the left and right audio channels, averaged over these videos.

- **Visual Salience Prior.** For a given test video clip, we consider all training set video clips made by the same YouTube channel. For each video, we use a horizontal sliding window of 32 pixels to determine the region of the videos with the largest squared distance to the average pixel value of the video. The prediction is given by the mode over these training clips.

- **Motion Prior.** For a given test video clip, we consider all training set video clips made by the same YouTube channel. For each video, we use a horizontal sliding window of 32 pixels to determine the region of the videos with the most motion, *i.e.*, largest squared distance between subsequent time frames. The prediction is given by the mode over these training clips.

As shown in Supplemental Table 1, these baselines fall short of the performance of the predicted ROI. Overall, our results strongly suggest that the flipping task trains the model to match the locations of sound sources in the visual frames to spatial audio cues, thus learning audio-visual spatial correspondence.

**Implementation Details for Sounding Face Tracking.** The architecture is shown in Figure 3(b) in the main text. The audio and visual sub-networks have the same architecture as those used for pretext task. The features from the sub-networks are stacked and passed through two deconvolution modules to expand the spatial grid size from 1-by-1 to 7-by-7, followed by convolution modules to ensure 40

| Feature | Pearson R | Spearman R |
|---|---|---|
| Sound prior | 0.062 | 0.101 |
| PCA proj | 0.709 | 0.790 |
| CCA proj | **0.909** | **0.876** |
| Salience prior | 0.008 | 0.012 |
| Motion prior | 0.009 | 0.001 |
| Predicted ROI [12] | **0.259** | **0.286** |

Table 1. Quantitative evaluation of different sound localization predictions based on Pearson (linear) correlation and Spearman (rank-based) correlation with the log energy-difference between left and right audio channels. Higher is better. The top results are for audio approaches and the bottom ones are for vision. The visual attention of our model outperforms the baselines.

outputs per spatial grid cell (4 coordinates and one confidence prediction for each of 8 anchor boxes). We obtain the anchor (prior) boxes using k-means clustering on the training data. We train our models on the YouTube-ASMR dataset, using 1-second audio clips sampled from full clips and a random visual frame outside of this range. The visual frame is resized to 256 x 256 and we shift the color and contrast as a form of data augmentation. For the audio input, we use the stacked log-transformed mel-spectrograms (number of frequency bins=512, window size=400 samples, hop size=160 samples) of the left and right audio channels sampled at 16 kHz computed using Librosa [7]. For optimization, we use Adam [5] with a learning rate of 1e-3, training on approximately 2M samples.

## C. Audio Spatialization (Upmixing)

**Implementation Details.** Our implementation for this task is based on [4]. The architecture is the U-Net shown in Figure 3(c) in the main text. The main difference is that we use multiple video frames to upmix a single audio clip, and we integrate these multiple frames into the U-Net by fusing the visual features with the reduced audio representation along the time dimension. We train and evaluate our models on both our YouTube-ASMR dataset and the FAIR-Play dataset. For both datasets, we use 2.87-second clips sampled from full clips. We use visual frames sampled at 6 Hz with frames resized to 256 x 256, and randomly crop and shift the color/contrast for data augmentation. For the audio input, we use the complex spectrogram (number of frequency bins=512, window size=400 samples, hop size=160 samples) of the difference between the left and right audio channels sampled at 16 kHz [7]. For optimization, we use Adam [5] with a learning rate of 1e-3, training on approximately 2M samples for YouTube-ASMR and 150K samples for FAIR-Play.

**Baselines.** All of the baselines use the same visual sub-network model architecture (ResNet-18) as our pretrained visual sub-network. For the audio-visual correspondence

task baselines, we train models on the tasks described in previous work (*i.e.*, [1] for audio-visual mismatch task and [9, 6] for the temporal shift task) over the same number of samples as our spatial correspondence task. The models have the same architecture as shown in Figure 3(a) in the main text, except they use average pooling rather than spatial flattening in the fusion of the video stream with the audio stream. For the supervised baseline, we use a ResNet-18 model pretrained on ImageNet classification [10].

### D. Audio-Visual Source Separation

**Implementation Details.** Our implementation for this task is based on [4]. We use the same model as the U-Net for audio spatialization, adapted for source separation (as shown in Figure 3(d) of the main text). We train and evaluate our models on both our YouTube-ASMR dataset and the FAIR-Play dataset. For both datasets, we use 2.87-second clips sampled from full clips. We use visual frames sampled at 6 Hz with frames resized to 256 x 256, and randomly crop and shift the color/contrast for data augmentation. For the audio input, we use the stacked magnitude spectrograms (number of frequency bins=512, window size=400 samples, hop size=160 samples) of the left and right audio channels sampled at 16 kHz [7]. For optimization, we use Adam [5] with a learning rate of 1e-3, training on approximately 2M samples for YouTube-ASMR and 200K samples for FAIR-Play.
**Baselines.** For the audio-visual correspondence task baselines, we train models on the tasks described in previous work (*i.e.*, [1] for audio-visual mismatch task and [9, 6] for the temporal shift task) over the same number of samples as our spatial correspondence task. The models have the same architecture as shown in Figure 3(a) in the main text, except they use average pooling rather than spatial flattening in the fusion of the video stream with the audio stream. Similar to the features for our audio-visual correspondence task, we use the joint audio-visual features after the fusion for the source separation task. For the supervised baseline, we use features from a ResNet-18 model pretrained on ImageNet classification [10].

### E. Spatial Alignment in 360-Degree Video

Here we discuss the implementation details and results of the audio-visual spatial correspondence task in 360-degree videos.

**Implementation Details.** We generally use the same model structure as the one we used for field-of-view video and stereo audio, as depicted in Figure 3(a) of the main text: the model consists of distinct visual and audio sub-networks that are fused prior to classification, and we reduce and flatten the visual features prior to fusion with the audio. The main difference is that the audio input is four channels instead of two. We train our model on the YouTube-

| Model | Test Accuracy |
|---|---|
| ResNet-18 (Supervised) | 0.57791 |
| No W-Channel | 0.61395 |
| No XYZ-Channels | 0.49535 |
| ResNet-18 | 0.58605 |

Table 2. Performance on pretext task: YouTube-360 dataset. The models perform comparably except the XYZ-channel ablation that eliminates spatial information.

360 dataset consisting of 360-degree videos with first-order ambisonics (FOA) audio [8]. The dataset consists of over a thousand 360-degree videos with corresponding spatial audio, containing a variety of content including street views and musical performances. We use 3 second video clips sampled from full-length videos, introducing negative (rotated audio) examples with probability 0.5 with $\theta \in [0.95\pi, 1.05\pi]$. We use 360-degree videos sampled at 5 Hz with frames (as equirectangular projection) resized to 240 x 480. To augment the dataset, we randomly shift the color/contrast of the videos and apply random rotation about the z-axis of both video (via translation of the equirectangular projection) and spatial audio. For the W-channel, we use the short-time Fourier transform (STFT) of audio sampled at 16 kHz, and for the other channels, we take the inner product of the channel STFT with the W-channel STFT, which is an effective input for spatial audio localization tasks using neural networks [3].

**Results.** Supplemental Table 2 shows the test classification accuracy of the spatial alignment task trained on YouTube-360 dataset. As a baseline, we initialized our visual sub-network using a model trained on ImageNet classification. We found that our model with visual sub-network trained from scratch performed comparably to this supervised baseline. Compared to earlier spatial alignment on the YouTube-ASMR and FAIR-Play datasets, these accuracy values appear low. We hypothesize that this is due to the in-the-wild nature of the dataset, resulting in sparser signals.

On the other hand, as shown in Supplemental Table 2, we find that a model with only 60% classification accuracy on 3-second clips aligns 360-degree videos with high accuracy when evaluated over the entire video (approximately 19 degrees of rotational error for high confidence videos). Specifically, for each video in the YouTube-360 test dataset, we compute the output of our model for uniform rotations of the spatial audio around the z-axis over multiple sub-sampled clips per video. We use consensus scoring to generate probabilities for different rotation angles and then compute the absolute value of the alignment error weighted by these probabilities. The results, specifically prediction errors in degrees, are shown in Table 3. As ablations/comparisons, we also trained the pretext task

|                  | Alignment Error (Degrees) |         |           |
|------------------|---------------------------|---------|-----------|
| Model            | Low Conf                  | Med Conf | High Conf |
| Original         | 65.34                     | 50.51   | 19.06     |
| Shorter clips    | 71.05                     | 55.23   | 38.21     |
| Harder negatives | 74.69                     | 66.65   | 47.00     |

Table 3. Evaluation of pretext models on audio-visual alignment in 360-degree videos, in degrees of error. Worst is 180 degrees, random is 90 degrees. Low conf shows the result averaged over all videos. Medium conf and high conf refer to approximately the 50th and 90th percentiles of videos based on the confidence of the model's classification, *i.e.*, predicted probability of a video being aligned correctly.

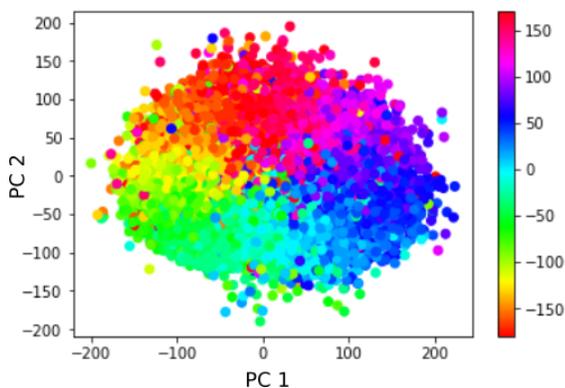

Figure 2. PCA visualization of audio embeddings extracted from Tau Spatial Sound dataset, colored by azimuth angle (in degrees). Each point corresponds to a different audio clip from the dataset.

on shorter clips (1 s) and harder negatives with smaller rotation angle of misaligned examples ($\theta \in [0.25\pi, 1.75\pi]$). We found that making the pretext detection task more challenging (*i.e.*, decreasing the clip length or increasing the difficulty of the negative examples) did not improve performance on this task.

**Qualitative analysis.** Qualitatively, we observe that the features learned by the pretrained audio sub-network capture the azimuth angle of the sound localization. Supplemental Figure 2 visualizes the first two principal components of audio embeddings extracted from the Tau Spatial Sound dataset, colored by the azimuth angle label. Notice that the angle of the first two principal components reflects the azimuth angle of the sound's direction of arrival. This observation enables us to perform acoustic localization using the learned audio embedding, *i.e.*, by overlaying the angle that is estimated based on the first two principal components over the equirectangular projection of the 360-degree video.


\end{document}

%% file: sections/introduction.tex
\section{Introduction}



\begin{figure}[t]
\begin{center}
\includegraphics[width=1\linewidth]{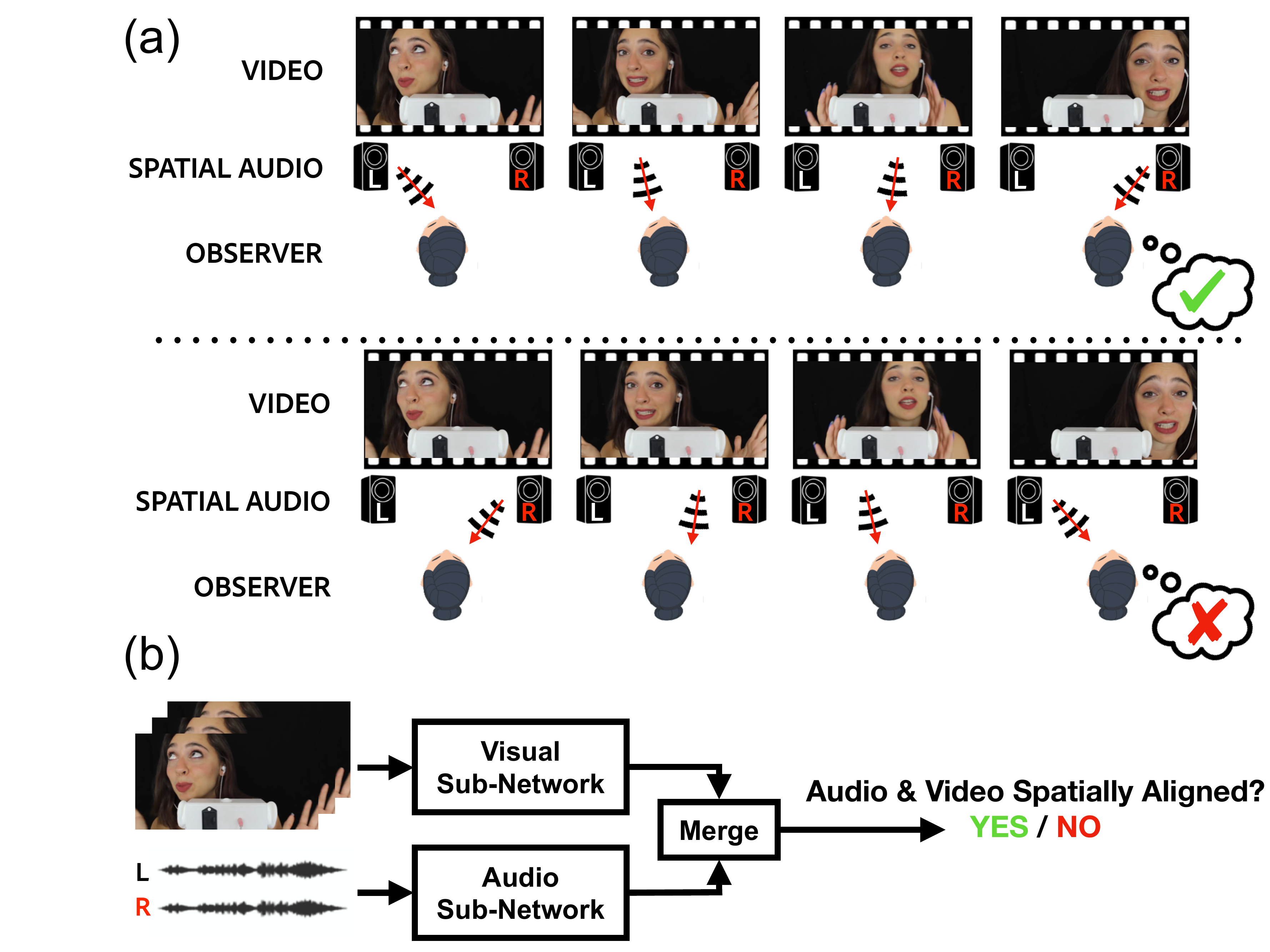} 
\end{center}
   \caption{(a) \REMOVED{Illustration of demo videos from Supplementary Material. }Based on the perceived location of the speaker using spatial audio cues, we can determine when the left/right positions of the sound sources based on sight and sound are aligned (top row) or flipped (bottom row). (b) We teach a model to understand audio-visual spatial correspondence by training it to classify whether a video's left-right audio channels have been flipped.
   }
\label{fig:intro}
\end{figure}

Consider Figure \ref{fig:intro}(a). Here, we illustrate two example videos with the perceived locations of a depicted speaker over time based on the spatial audio%
\footnote{These videos are provided in the Supplementary Materials. We encourage you to watch and listen to the videos wearing headphones.}. 
In the first video, notice that our spatial perception of the speaker moving from left to right is consistent between the visual and auditory streams, while in the second video, there is an obvious discrepancy between the two modalities (the left and right audio channels have been flipped, so the sound comes from the wrong direction). This effect is due to the spatial audio signal in these two videos: the audio emulates our real-world auditory experience by using separate left and right (\ie, stereo) channels to deliver binaural cues influencing spatial perception \cite{rayleigh1875our}. Because humans have the ability to establish spatial correspondences between our visual and auditory senses, we can immediately notice that the visual and audio streams are consistent in the first video and flipped in the second video. Our ability to establish audio-visual spatial correspondences enables us to interpret and navigate the world more effectively %
(\eg, a loud clatter draws our visual attention telling us where to look; when interacting with a group of people, we leverage spatial cues to help us disambiguate different speakers). In turn, understanding audio-visual spatial correspondence could enable machines to interact more seamlessly with the real world, improving performance on audio-visual tasks such as video understanding and robot navigation. 


Learning useful representations over visual and audio streams in video is challenging. While strong features have been learned using strongly supervised training data \cite{hershey2017cnn}, large quantities of annotations are difficult to obtain. To overcome this challenge, many self-supervised learning approaches have recently been proposed to exploit the audio-visual correspondence in video as a free source of labels \cite{Arandjelovi17, Ephrat18, Gao18,Gao19b, Korbar18,Owens18,Rouditchenko19, Zhao19, Zhao18}. These approaches learn audio and visual representations primarily by matching semantic information or temporal information in audio 
to the presence or motion of sound sources in the visual stream, without leveraging the audio-visual spatial relation. In contrast, we seek to explicitly focus on spatial correspondence and explore a completely orthogonal approach to audio-visual feature learning based on matching spatial cues in the audio stream to the positions of sound sources in the visual stream. At a time when videos with spatial audio are rapidly proliferating (\eg, due to advances in cellphone mics and AR/VR technology), understanding how to leverage these data to learn strong audio and visual representations is of significant scientific and practical interest. 

In this work, we investigate a simple, yet effective, way to teach machines to understand audio-visual spatial correspondence -- learn to classify whether a video's left-right audio channels have been flipped, as illustrated in Figure 1(b). We conjecture that a model needs to establish spatial correspondence between audio and visual inputs in order to solve this task. 
The primary contribution of our work, therefore, is to study the extent to which spatial understanding is useful by (i) proposing a novel self-supervised pretext task for teaching audio-visual spatial correspondence and (ii) evaluating the learned features on an array of downstream audio-visual tasks that could potentially benefit from a strong multimodal spatial representation. Critical to the evaluation of our task is the ability to train on a large video dataset with spatial audio. As part of our contribution, we introduce a new video dataset of ASMR videos from YouTube (``YouTube-ASMR-300K"), the largest reported video dataset with spatial audio comprising over 900 hours of footage. We demonstrate that machines improve on audio-visual tasks through spatial understanding. Over three downstream tasks -- (1) sound localization, (2) audio spatialization (upmixing a single mono audio channel to stereo binaural audio channels), and (3) sound source separation -- we achieve a quantitative improvement over prior self-supervised audio-visual correspondence tasks and over strongly supervised visual features alone. \ADDED{We also extend the left-right pretext task to 360 degree videos with ambisonic audio and apply the learned embeddings to 360 degree sound localization.}

%% file: sections/related.tex
\section{Related Work}

\paragraph{Self-supervised learning in videos.}
 Due to the challenges in obtaining large-scale annotated data for supervised training, many works have proposed \REMOVED{approaches }leveraging audio-visual correspondences for self-supervised learning \cite{Arandjelovi17, de1994learning,Korbar18, ngiam2011multimodal,Owens18, owens2016visually, owens2016ambient}. These tasks learn audio-visual representations either by leveraging the shared semantic information \cite{Arandjelovi17, owens2016ambient} or by exploiting temporal correlation \cite{Korbar18, ngiam2011multimodal, Owens18, owens2016visually}. However, they do not exploit spatial correspondence for self-supervision, and most only take mono audio as input. 

Inspiring our effort is prior work on audio-visual correspondence tasks for predicting whether visual and audio signals come from the same video \cite{Arandjelovi17}, and whether visual and audio signals are temporally aligned \cite{Korbar18, Owens18}. These tasks learn audio and visual representations based on correspondence in semantic \REMOVED{information }or temporal information. In contrast, our \REMOVED{audio-visual} correspondence task is designed to teach a model to match spatial audio cues to positions of sound sources in the video and \REMOVED{focuses on exploiting}\ADDED{exploits} the spatial relation.

\paragraph{Audio-visual source separation.}
Audio-visual source separation utilizes visual information to aid in the separation of sound mixtures. Many self-supervised approaches have been proposed to solve this task \cite{Ephrat18,Gao18,Gao19b, Owens18, Rouditchenko19, Zhao19}, including mix-and-separate frameworks that combine audio tracks from videos and then train models to separate them using visual information \cite{Ephrat18, Owens18, Zhao19,Zhao18}. In order to leverage the visual cues, these separation models learn an audio representation that captures the semantics \cite{Zhao18} or temporal patterns \cite{Ephrat18,Zhao19} of sound in order to match them to the visual frames respectively. Strategies to separate audio without explicit mixtures \cite{Gao18,Gao19b} or to co-segment video at the same time \cite{Rouditchenko19} have also been proposed. However, all of these approaches still learn an audio representation based on matching semantics or temporal patterns and do not leverage spatial cues.

\paragraph{Sound localization.}
Estimating the direction-of-arrival of a sound using multiple microphones has been traditionally tackled using beamforming algorithms such as steered power response \cite{butko2011two} that do not learn an audio representation and cannot easily handle the concurrency of multiple sound sources. 
More recently, methods based on neural networks have been proposed for direction of arrival estimation \cite{adavanne2018sound,hirvonen2015classification,lopatka2016detection}. These models learn spatial audio representations, but they are trained through strong supervision, while we propose to use self-supervision to learn spatial audio cues by leveraging audio-visual spatial correspondence. 

A separate stream of research has focused on localizing sound sources in videos \cite{barzelay2007harmony, fisher2001learning,  hershey2000audio, kidron2005pixels}, including in self-supervised ways \cite{arandjelovic2018objects, Owens18,senocak2018learning}. However, these methods do not exploit spatial audio. Recently, Gan \etal \cite{Gan19} proposed learning a spatial audio representation that localizes vehicles in video \REMOVED{using camera metadata as ground truth for training}\ADDED{using a pretrained vision network as a teacher}. In contrast, our \REMOVED{spatial audio }representation is learned \REMOVED{in a more general setting, }by leveraging audio-visual spatial correspondence without explicitly modeling the target locations \REMOVED{in the video}\ADDED{of a teacher}.

\paragraph{Audio spatialization.}
Several self-supervised approaches have recently been proposed for audio spatialization, which is the task of converting mono audio to spatial audio using a concurrent visual stream to inject the audio with spatial cues \cite{Gao19, Morgado18, Lu19}. Similar to us, these approaches use spatial audio as a self-supervisory signal. For example, Gao \etal~\cite{Gao19} propose a U-Net architecture with a visual stream to convert a mono audio input (generated by downmixing stereo audio) to a stereo output, using the original stereo audio as the target during training. In their model, the visual features provide complementary spatial information that is missing from the mono audio to produce stereo audio. In contrast, our audio representation learns spatial cues directly from stereo audio in order to match the perceived localization of a sound with its position in the video. Also similar to us, Lu \etal~\cite{Lu19} proposed a spatial correspondence classifier, but they apply it as an adversarial loss for aiding audio spatialization whereas we propose the spatial audio-visual correspondence task for self-supervised feature learning.

%% file: sections/stereo.tex
\section{\asmr~Dataset} \label{sec:asmr-dataset}

\begin{figure}[t]
\begin{center}
   \includegraphics[width=1\linewidth]{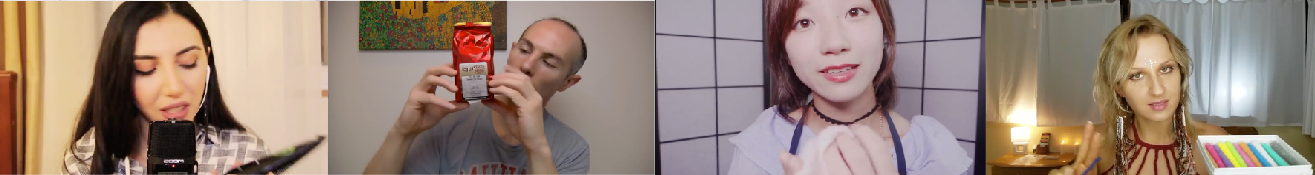}
\end{center}
   \caption{Examples of videos from our \asmr~ dataset.}
\label{fig:ASMR-examples}
\end{figure}

Learning from videos with spatial audio is a relatively new domain. While the amount of spatial audio content is \REMOVED{rapidly proliferating}\ADDED{increasing}, currently there are few video datasets with spatial audio in which the visual content is spatially aligned with the audio content. We therefore introduce a new large-scale dataset of ASMR videos collected from YouTube that contains stereo audio. ASMR (autonomous sensory meridian response) videos \ADDED{are readily available online and} typically feature an individual actor or ``ASMRtist" making different sounds while facing towards a camera set up with stereo/binaural or paired microphones. Screenshots from our dataset \REMOVED{is}\ADDED{are} shown in Figure \ref{fig:ASMR-examples}. The \REMOVED{stereo }audio in these videos contain\ADDED{s} binaural cues that, when listened to with headphones, create a highly immersive experience in which listeners perceive the sounds as if they were happening around them. Thus there is strong correspondence between the visual and spatial audio content in these videos. 

Our full dataset, \REMOVED{\asmr}\ADDED{YouTube-ASMR-300K}, consists of approximately 300K 10-second video clips with spatial audio. From this full dataset, we also manually curated a subset of 30K clips from 30 ASMR channels that feature more sound events moving spatially for training our models. We call this curated dataset \asmr. We split the video clips into training, validation, and test sets in an $80$-$10$-$10$ ratio. \REMOVED{As shown in Table \ref{table:datasets}, \asmr~is 8X larger than the largest prior dataset with spatial audio, opening the opportunity to learn audio-visual features at scale.}\ADDED{Compared to the existing datasets, YouTube-ASMR-300K is (1) larger by at least 8X (Table \ref{table:datasets}), (2) collected in-the-wild, and (3) contains sound sources in motion (\eg, a user waves a tuning fork across the field of view), which is important for training models on diverse spatial cues.} \REMOVED{Upon publication of the work, the video URLs and splits for this dataset will be made publicly available.}\ADDED{The dataset URLs are available on the project website listed on the title page.}

\begin{table}[!t]
\begin{center}
\resizebox{\linewidth}{!}{
\begin{tabular}{|l|c|c|}
\hline
Dataset & \# Unique Videos & Duration (Hrs) \\
\hline
Lu \etal \cite{Lu19} & N/R & 9.3 \\
FAIR-Play \cite{Gao19} & N/R $^*$ & 5.2 \\
YouTube-360 \cite{Morgado18} & 1146 & 114 \\
\hline
\asmr & 3520 & \REMOVED{97}\ADDED{96} \\
YouTube-ASMR-300K & \bf{33725} & \REMOVED{\bf{924}}\ADDED{\bf{904}} \\

\hline
\end{tabular}
}
\end{center}
\caption{Current large-scale video datasets with spatial audio. Our YouTube-ASMR-300K dataset has the most unique clips and the longest total duration. $^*$2000 10-second clips in total.}
\label{table:datasets}
\end{table}

\section{Learning to Tell Left from Right} \label{sec:stereo-pretext}

\begin{figure*}[t]
\begin{center}
\includegraphics[scale=0.09]{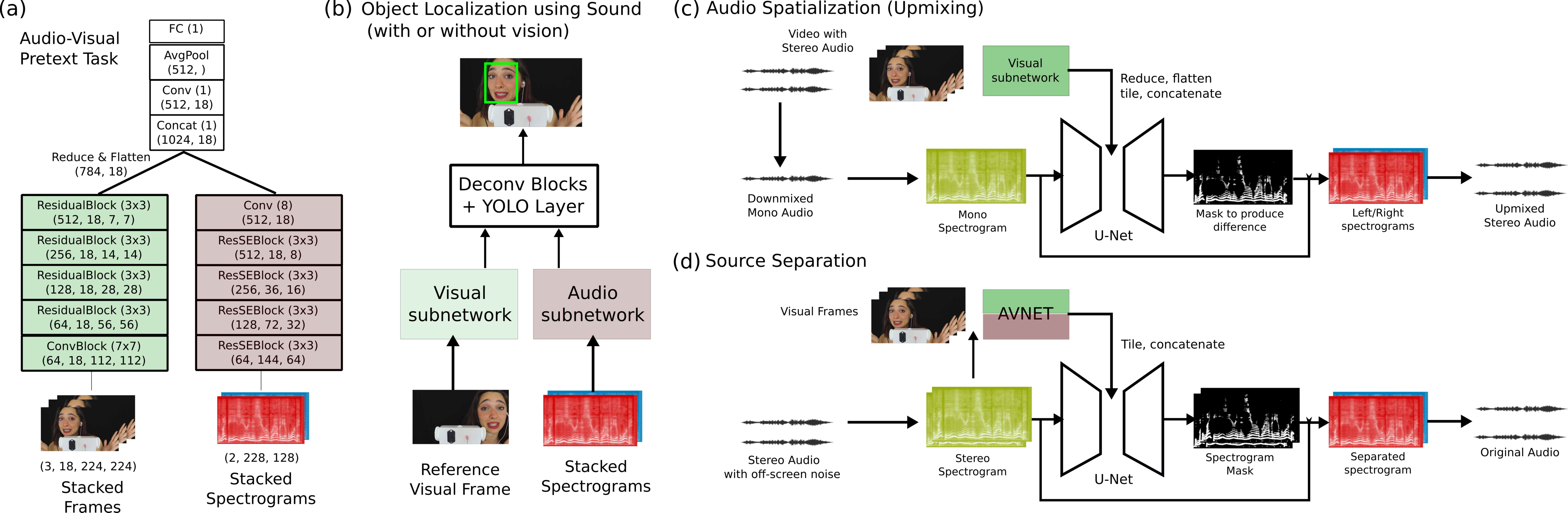}
\end{center}
\caption{(a) The model architecture for our spatial audio-visual correspondence task. See main text for details. (b-d) Network architectures leveraging our pretrained features for downstream tasks: (b) \REMOVED{sound localization}\ADDED{localization/tracking}, (c) audio spatialization, and (d) source separation. }
\label{fig:architectures}
\end{figure*}

\paragraph{Problem Formulation.} Spatial audio enables listeners to infer the locations of sound sources. In the case of stereo audio, binaural cues such as differences in time of left and right signals arriving (inter-aural time differences) and the difference in levels of the left and right signals (inter-aural level differences) contribute to the perception of sound sources being localized to the left or right \cite{rayleigh1875our}. We hypothesize that these binaural cues can be used to learn useful multimodal representations by teaching a model to recognize when the video and audio are not spatially aligned. 
During training, we provide as input video clips where we flip the order of the channels in the audio stream with probability 0.5, \ie, if the original audio is given by $a(t) = (a_l(t), a_r(t))$, where $a_l(t)$ and $a_r(t)$ are the left and right channels as a function of time $t$, the flipped audio is $\tilde{a}(t) = (a_r(t), a_l(t))$. This transformation switches the inter-aural time and level differences between the \REMOVED{left and right }audio channels. 

Formally, let $D= \{(v,a,y)_i\}_{i=1}^N$ be our video dataset where $v$ is a visual stream, $a$ is an audio stream, and $y$ indicates whether or not $a$ is flipped with respect to $v$. We train the neural network $f_w(v,a)$ with parameters $w$ to maximize a classification cross-entropy objective given by the log-likelihood,
\begin{align}
\sum_{(v,a,y) \in D} y \log f_w(v,a) + (1-y) \log (1-f_w(v,a)).
\end{align}
We conjecture that solving the flipping task requires understanding audio-visual spatial correspondence, as the model must match the location of objects in audio signals with the location of objects visual signals.

%
\paragraph{Spatial Alignment Network.} We illustrate our network in Figure \ref{fig:architectures}(a). The network comprises two streams -- a visual stream and an audio stream -- which are fused \REMOVED{together in}\ADDED{by concatenating features along} the time dimension, followed by additional convolutional layers to yield an output representation prior to classification. For vision, our base model uses the public PyTorch implementation of ResNet-18 \cite{he2016deep} (note that we only apply spatial and not temporal convolutions). We use frames sampled at 6 Hz and resized to 256 x 256 as input. For audio, our base model uses stacked residual blocks with S\&E \cite{hu2018squeeze}, matching the output temporal dimension to that of the vision network. We use the log-scaled mel-spectrogram of audio sampled at 16 kHz and stack the left and right stereo channels as input. Our two-stream network is comparable to the architectures of previous audio-visual correspondence tasks \cite{Arandjelovi17, Korbar18}. However, a key distinction is that our model requires the positions of sound sources \REMOVED{learned}\ADDED{detected} by the visual sub-network. While spatial pooling of the features from the visual sub-network prior to fusion is suitable for \REMOVED{previous}\ADDED{other} correspondence tasks, \REMOVED{a necessary design choice for this task is to flatten the visual features}\ADDED{we found it necessary for our task to flatten the visual features along the spatial dimensions} without pooling prior to fusion with the audio. Models used for audio spatialization tasks, which also require knowledge of the position of sound sources in the visual frame, process visual features in a similar way \cite{Gao19,Morgado18}. For applications to downstream tasks, we can use features from the audio sub-network, visual sub-network, or the fused representation. 

\paragraph{Training.}
We train and evaluate our models on both our \asmr~dataset and the FAIR-Play dataset \cite{Gao19}. The latter dataset consists of approximately 2K 10-second video clips of people playing instruments. While this dataset is smaller than \asmr, we use it to demonstrate the generality of our approach. For both datasets, we use 3-second clips sampled from full clips, introducing flipped audio examples with probability 0.5. We apply a random crop and shift the color/contrast of the frames for data augmentation. To account for possible left-right biases in the audio and visual information, we apply random left-right flipping of both \ADDED{the }video and audio channels (\REMOVED{note: flipping both at the same time does not affect the prediction target of whether or not the video and audio are spatially aligned}\ADDED{note: flipping both at the same time maintains the audio-visual spatial alignment or lack thereof, and thus does not change the prediction target}). For optimization, we used SGD (momentum=0.9) with a learning rate of 0.01 on up to 7M samples for YouTube-ASMR and 800K samples for FAIR-Play using 1-4 GPUs.

\paragraph{Baselines and Ablations.}
Our base model is trained from scratch using ResNet-18 as the visual sub-network architecture as shown in Figure \ref{fig:architectures}(a). To determine if our model can \REMOVED{optimize visual features for this task in a self-supervised manner}\ADDED{obtain effective visual features with self supervision}, we compared against a baseline using ResNet-18 pretrained on ImageNet classification and finetuned on our task (``Pretrained on ImageNet"). To assess the importance of motion features, we trained a model from scratch using MCx as the visual sub-network \cite{tran2018closer}, which uses 3D spatiotemporal convolutions and is designed for video classification (``Motion"). To determine if \REMOVED{the task performance is improved using semantic audio features}\ADDED{semantic audio features improve the task performance}, we force the model to learn semantic audio features in a third separate stream that only takes mono audio input (``+Mono audio"). \ADDED{This third stream uses the same audio sub-network as stereo audio (with one input channel instead of two) and is similarly fused by concatenating features along the time dimension.} Finally, to determine whether the learned audio features additionally benefit from having traditionally computed spatial audio cues, we integrated features computed using Generalized Cross Correlation with Phase Transform \cite{Cao2019} to the audio stream \REMOVED{in the dual audio model}\ADDED{by introducing three additional channels in the stereo audio input} (``+GCC-Phat").

\paragraph{Results.}
\REMOVED{The test set classification accuracy of the audio-visual spatial correspondence model trained on \asmr~and FAIR-Play is reported}\ADDED{We report the test set classification accuracy of the audio-visual spatial correspondence model trained on \asmr~and FAIR-Play} in Table \ref{table:pretext-asmr}. \REMOVED{The base}\ADDED{Our} models perform well and comparably to the supervised baseline; in fact, the performance on the \asmr~dataset is comparable to human performance on a sub-sample of 200 clips (about $80\%$, N=2 subjects). This result indicates that our model architecture is well-suited for matching spatial audio cues with sound source locations in the video frames, and also that the spatial audio cues in both datasets are sufficiently rich for learning the visual features of sound sources. We did not obtain gains using the MCx network, suggesting that 3D spatiotemporal convolutional features may not be integral to the task on these datasets. We observed gains using the dual audio model (``+Mono Audio"), indicating that using semantic audio features could potentially aid performance on the pretext task (\eg, when there are multiple sound sources coming from different directions). Finally, GCC-Phat features did not improve the model performance, which suggests that our audio sub-network learned, in a fully self-supervised manner, the spatial localization cues that are traditionally computed using beamforming algorithms. In the downstream task analysis, we use our base model with the ResNet-18 visual sub-network trained from scratch to focus on evaluating audio-visual features learned using spatial audio cues rather than semantic cues.

\begin{table}[!t]
\begin{center}
{
\begin{tabular}{|l|c|c|}

\hline
Model & \asmr & FAIR-Play \\
\hline
Pretrained on ImageNet & 80.4 & 92.6 \\
\hline
\REMOVED{Base}\ADDED{Ours} & 80.1 & 93.6 \\
\hline
Motion & 80.4 & 71.3 \\
\REMOVED{Dual audio}\ADDED{+Mono Audio} & \textbf{81.3} & \textbf{96.3} \\
+GCC-Phat & 80.1 & 94.1 \\
\hline
\end{tabular}
}
\end{center}
\caption{Test set classification accuracy of the pretext task trained on the \asmr~ and FAIR-Play datasets. Our base models trained from scratch perform comparably or outperform a model that uses supervised features from ImageNet classification.}
\label{table:pretext-asmr}
\end{table}

\section{Analysis on Downstream Tasks}

\subsection{Sound Localization} \label{sec:sound-source-loc}

\begin{figure*}[ht]
\begin{center}
\includegraphics[scale=.05]{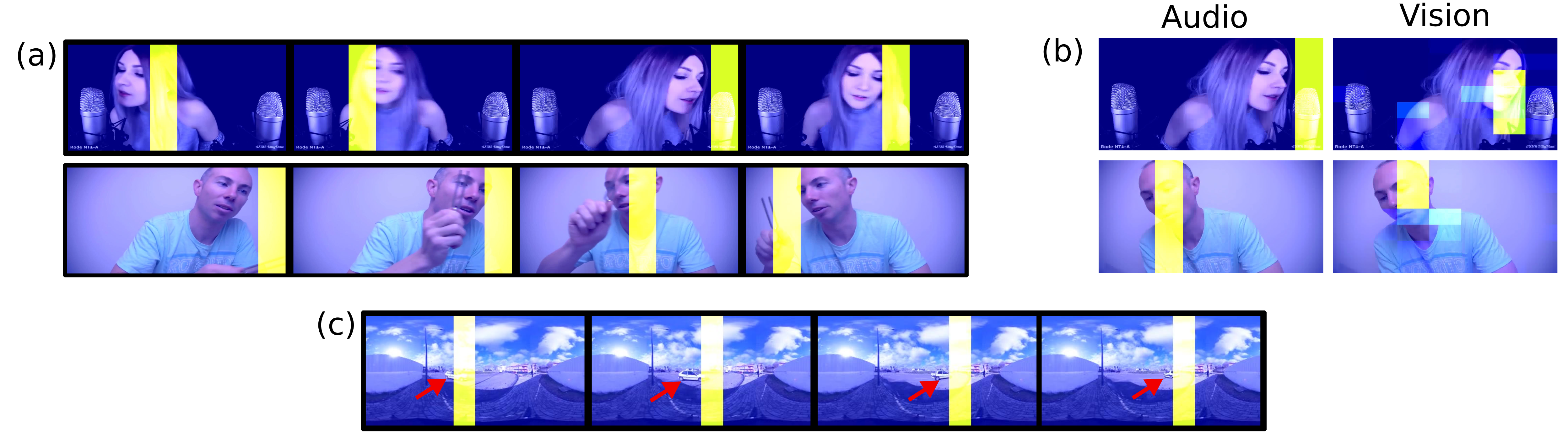}
\end{center}
   \caption{(a) Sound localization results on \asmr~using audio only. \REMOVED{The visualization is generated by binning the values of the first principal component of the learned audio embedding over the range of the video.} (b) Comparison of localization results using the spatial cues captured by the audio sub-network (left) and the regions of importance of the visual sub-network (right). The visual attention of the model is localized to sound sources and corresponds to the spatial cues learned by the audio sub-network. \ADDED{(c) Sound localization results on YouTube-360 using audio only.}}
   \label{fig:asmr-audio-visual-loc}

\end{figure*}

\REMOVED{Does the audio-visual spatial correspondence task teach our model to match spatial audio cues to the positions of sound sources in the visual stream?}\ADDED{Does our spatial correspondence task learn an effective representation for mapping spatial audio cues to the positions of sound sources in the visual stream?} To answer this question, we first evaluate whether audio embeddings extracted from held-out data contain spatial information. \ADDED{Specifically, we compute the correlation between learned audio features and the sound source's approximate spatial location, which we determine based on the log-energy difference between the two audio channels. We find that the audio features are strongly correlated with the location of the sound source ($R=0.790$, see Supplemental)}. \REMOVED{We achieve this goal by comparing the learned audio features to a sound source's spatial location at each time instance. To obtain an approximate location of a sound source at a time instance, we compute the log-energy difference between the two audio channels. In Table \ref{table:sound-loc}, we report Pearson R and Spearman R correlation coefficients to measure the correlation between the approximate sound location computed from directional energy and the self-supervised learned embedding. We find that the audio features are strongly correlated with the location of the sound source based on both an unsupervised projection (\ie, principal component analysis) and a supervised projection (\ie, canonical correlation analysis) of the learned embeddings.}
In Figure \ref{fig:asmr-audio-visual-loc}(a), we show sound sources automatically tracked over time using our audio embeddings, visualized by a vertical yellow bar in each frame. \ADDED{We generate the visualizations by binning the values of the first principal component of the learned audio embedding and assigning the bins to different horizontal locations in a video frame.} As an example, notice in the first row that the yellow bar follows the sound of the moving subject from \REMOVED{right to left}\ADDED{left to right}. These results suggest that the audio sub-network learns spatial cues in order to \REMOVED{perform}\ADDED{solve} the proposed audio-visual correspondence task. 

\REMOVED{
\begin{table}[!t]
\begin{center}
\begin{tabular}{|c|c|c|}
\hline
Feature & Pearson R & Spearman R \\
\hline
Sound prior & 0.062 & 0.101 \\
PCA proj & 0.709 & 0.790 \\
CCA proj & 0.909 & 0.876 \\
\hline
Salience prior & 0.008 & 0.012\\
Motion prior & 0.009 & 0.001 \\
Predicted ROI \cite{zeiler2014visualizing} & \textbf{0.259} & \textbf{0.286} \\
\hline
\end{tabular}
\end{center}
\caption{Quantitative evaluation of different sound localization predictions based on Pearson (linear) correlation and Spearman (rank-based) correlation with the log energy-difference between left and right audio channels. Higher is better. The top results are for audio approaches and the bottom ones are for vision. The visual attention of our model outperforms the baselines. }
\label{table:sound-loc}
\end{table}
}

\REMOVED{Next, we investigated whether the visual sub-network has learned to identify the positions of sound sources, as a result of having to match them with the spatial cues provided by the audio sub-network.}\ADDED{Next, we evaluate whether our model has learned to match the spatial cues in the audio embedding to the positions of sound sources in the visual stream. Specifically, we determine to which regions the visual sub-network attends as described in the Supplemental. We find \REMOVED{using this method}\ADDED{qualitatively} that visual attention is \REMOVED{convincingly }directed to the sound sources in the visual frame, and that there is notable correlation between the region of visual attention and the approximate location of a sound source based on the log-energy difference between the two audio channels ($R=0.286$, Supplemental).}%
\REMOVED{To determine the regions that the visual sub-network attends to, we use a sliding window of $32$ pixels and replace regions of the visual frame with their mean values. We then pass the modified frames through our visual sub-network and determine those regions whose omission maximally affects prediction on the pretext task \cite{zeiler2014visualizing, Gao19}. We find using this method that visual attention is convincingly directed to the sound sources in the visual frame, and that there is significant correspondence between regions of visual attention and spatial cues learned by the audio sub-network.} In Figure \ref{fig:asmr-audio-visual-loc}(b), we have visualized examples of this correspondence. Notice in the top row that the visual sub-network attends to the speaker's face (right image), in correspondence with the audio sub-network, which identifies the sound as coming from the right (left image). \ADDED{Overall, the analysis suggests that our spatial alignment model learns a representation that maps spatial audio cues to the positions of sound sources in the visual stream.}
\REMOVED{The spatial audio in the \asmr~dataset provides a unique opportunity for us to evaluate the visual attention results in a quantitative manner. Table \ref{table:sound-loc} shows our evaluations of the predicted regions of importance using the masking approach of \cite{zeiler2014visualizing, Gao19} (``Predicted ROI") by comparing them to the approximate ground truth locations. We find that the predicted localization results are significantly correlated with the approximate ground truth locations. To preclude the possibility that the visual network is relying on artist-dependent spatial biases to achieve this correlation rather than the dynamic spatial audio cues, we generate baseline predictions of sound localization based on the prior distribution of the sound, visual salience, and motion in training videos produced by the same artists using metadata information (details are provided in the Supplemental). These baselines fall short of the performance of the predicted ROI. Overall, our results strongly suggest that the flipping task trains the model to match the locations of sound sources in the visual frames to spatial audio cues, thus learning audio-visual spatial correspondence.}

\begin{table}[!t]
\begin{center}
\begin{tabular}{|c|c|c|c|c|}
\hline
Model & AP50 & IOU & X-error & Y-Error \\
\hline
Mono audio & 10.2 & 30.6 & 11.9 & 10.4 \\
Stereo audio & 15.1 & 34.8 & 10.0 & 10.7 \\
Visual context & 27.9 & 41.0 & 12.9 & \bf{5.5} \\
Ours & \bf{43.4} & \bf{47.1} & \bf{9.1} & \bf{5.5} \\
\hline
\hline
Mono audio & 13.9 & 23.8 & 17.3 & 10.0 \\
Stereo audio & 26.0 & 39.0 & \bf{9.7} & 9.6 \\
Visual context & 21.6 & 27.0 & 20.7 & 6.5 \\
Ours & \bf{44.1} & \bf{45.0} & 10.8 & \bf{5.9} \\
\hline
\hline
- Pretext & 23.9 & 35.1 & 12.4 & 8.9 \\
+ Pretext & \bf{34.9} & \bf{39.5} & \bf{12.3} & \bf{6.5} \\
\hline
\end{tabular}
\end{center}
\caption{Results of audio-visual localization on the \asmr~ test data (Rows 1-4, 9-10) and filtered test data (Rows 5-8). See text for details. Our model outperforms the baselines, and using the pretrained weights from the flipping pretext task outperforms no pretraining.}
\label{table:loc1}
\end{table}

\paragraph{Tracking Sounding Faces using Stereo Sound.} \ADDED{To leverage our pretrained features for sound localization, we devise a new audio-visual face tracking task on the \asmr~dataset. The goal is to generate bounding boxes for sounding faces using stereo audio and a visual reference frame taken from a different part of the same video (Figure \ref{fig:architectures}(b)). In practice, such a system could be used to augment vision-based tracking, \eg, spatial sound can allow a system to reason through visual occlusion of a sounding object. Although the sounds in \asmr~are produced by a variety of objects, we focus on tracking faces for three reasons: (1) almost all videos feature front-facing individuals, (2) many sounds are mouth-based sounds (\eg, whispering, eating) that contain useful signal for localizing faces, and (3) pretrained vision networks such as RetinaFace \cite{deng2019retinaface} provide a cheap yet reliable source of pseudo ground-truth labels. Our task is motivated by Gan \etal \cite{Gan19}, which proposes tracking vehicles from stereo sound and camera metadata without visual input.}

\paragraph{\ADDED{Model.}} 
\ADDED{Our model (shown in Figure \ref{fig:architectures}(b)) is built around the pretrained audio and visual sub-networks from our pretext task. For input, the model takes a one-second audio clip centered around the frame containing the localization target, as well as a reference visual frame from a different part of the same video. The inputs are passed through the sub-networks and concatenated in a one-dimensional vector. Similar to Gan \etal \cite{Gan19}, the features are then passed through several deconvolution blocks to predict the coordinates of bounding boxes relative to anchor (prior) boxes. We use the object detection loss of YOLOv2 \cite{redmon2017yolo9000} for training. We initialize the audio and visual sub-networks with pretrained weights from the pretext task and fine tune through the entire network.}

\paragraph{\ADDED{Results.}} 
\ADDED{For evaluation, we consider three metrics: (1) average precision with intersection-over-union (IOU) threshold set at 50 (``AP50"), (2) average IOU of the highest-confidence box per frame (``IOU"), and (3) average error in the x and y coordinates of the highest-confidence box per frame as a percentage of the frame dimension (``X-error", ``Y-error"). Rows 1-4 of Table \ref{table:loc1} show the performance of our model against several baselines: models that only use mono or stereo audio (``mono audio" and ``stereo audio"), and a model that only uses visual context (``visual context"). We find that our model outperforms all of the baselines. However, it is surprising that the performance boost from using stereo audio is not larger. We hypothesize that many videos may feature stationary faces (enabling the visual context model to do well), or contain sounds that do not originate from the face. Therefore, we also evaluate our models on a subset of test videos that are likely to contain moving faces and mouth sounds, \ie, we filtered for videos in which there are large left-right shifts in the face's horizontal location that correlate with changes in log-energy difference between the two audio channels. Rows 5-8 of Table \ref{table:loc1} show the models evaluated on this subset of test clips. We now observe a significant improvement when using stereo audio. Finally, to determine the extent to which our pretrained features from the pretext task are helpful for this task, we train our model on the full data with fixed sub-network weights (no fine-tuning), with and without pretraining. Using the pretrained weights  (``+ pretext task") improves performance across all metrics compared to no pretraining (``- pretext task") (Table \ref{table:loc1}, Rows 9-10).}


\subsection{Audio Spatialization (Upmixing)} \label{sec:upmixing}
The goal of \REMOVED{audio spatialization}\ADDED{upmixing} is to convert \REMOVED{a single audio channel (mono audio)}\ADDED{mono audio} to multi-channeled spatial audio, providing the listener with the sensation that sounds are localized in space. Recent work has used a concurrent visual stream to provide spatial information \cite{Gao19, Lu19, Morgado18}. Specifically, the model is tasked with upmixing the audio stream by matching sounds to the perceived locations of their sources in the \REMOVED{corresponding }video frame. 
Since our \ADDED{spatial alignment} model matches spatial audio cues to visual sound sources, we hypothesize that the pretrained \REMOVED{visual }features could be useful \REMOVED{towards audio spatialization}\ADDED{for the upmixing task}.

\paragraph{Model.} We adopt the Mono2Binaural framework of Gao \etal \cite{Gao19}. The model takes mono audio and visual frames as input and produces binaural (two-channel) spatial audio as output by producing a complex mask for the difference between the two channels, illustrated in Figure \ref{fig:architectures}(c). We use a U-Net to upmix the audio input, and \REMOVED{inject visual information by concatenating extracted}\ADDED{concatenate our pretrained} visual features to the innermost layer of this network. Our model implementation is almost identical \ADDED{to that of Gao \etal \cite{Gao19}}, except we use Tanh activation for producing the complex mask instead of Sigmoid activation\REMOVED{ layer}\REMOVED{. In our early experiments directly applying \REMOVED{the Mono2Binaural}\ADDED{their} model to \asmr, we noticed qualitatively that the upmixing was biased in one direction due to the asymmetry of the Sigmoid layer for producing the difference mask}\ADDED{ because we noticed that the asymmetry of the Sigmoid layer for producing the difference mask biases the upmixing in one direction}. These effects were only noticeable because of the strong binaural cues in our dataset. Switching to a Tanh activation layer resolved the bias and quantitatively improved the results, so we maintained this change.

\paragraph{Baselines and Evaluation Criteria.} We compare our pretrained visual sub-network features (``flip task") against several baselines: (i) No visual features; (ii) ResNet-18 without any training (``no pretraining"); (iii) ResNet-18 trained on the audio-visual correspondence tasks of detecting mismatching semantic information (``mismatch task") \cite{Arandjelovi17} or (iv) shifted temporal alignment (``shift task") \cite{Korbar18,Owens18}, which use non-spatial audio cues to learn visual features; (v) ResNet-18 trained on ImageNet (``supervised"), which is the visual network used in Gao \etal \cite{Gao19}. \ADDED{For training and evaluation criteria, we used the L1 distance between the output and target complex spectrograms averaged over time-frequency bins.}

\paragraph{Results.} The test set errors are shown in Table \ref{table:upmixing}. Our pretrained features from the spatial alignment detection task \REMOVED{provide a considerable boost to}\ADDED{improve} audio spatialization on both the \asmr~and the FAIR-Play datasets, outperforming all of the baselines. \ADDED{The difference between our model and the rest is significant based on Wilcoxon signed-rank tests on \asmr~ ($p < 0.1$ for shift task, $p < 0.05$ for rest)\footnote{See project website for spatialization examples}.} This result suggests that \REMOVED{the spatial alignment detection}\ADDED{our pretext} task, which uses spatial audio cues to guide the video sub-network, successfully teaches the visual sub-network to extract features corresponding to sound sources. Importantly, our features outperform the pretrained features from other audio-visual self-supervised correspondence tasks that use non-spatial cues trained on the same dataset. These results suggest that in the YouTube-ASMR and FAIR-Play datasets, the spatial information in the audio may be a richer source of signal than the semantic information. Overall, we show that spatial audio cues \REMOVED{can serve as}\ADDED{are} a powerful alternative to semantic audio cues for learning visual features in a self-supervised manner. 

\begin{table}[!t]
\begin{center}
\begin{tabular}{|c|c|c|}
\hline
Visual sub-network & \asmr & FAIR-Play \\
\hline
Supervised & 0.0858 & 0.403 \\
\hline
No visual features & 0.0924 & 0.418 \\
No pretraining & 0.0891 & 0.413 \\
\hline
Mismatch task \cite{Arandjelovi17} & 0.0877 & 0.412 \\
Shift task \cite{Korbar18,Owens18} & 0.0861 & 0.409 \\
Flip task (ours) & \textbf{0.0853} & \textbf{0.401} \\
\hline
\end{tabular}
\end{center}
\caption{Test set error of upmixing on the \asmr~ and FAIR-Play datasets. Our pretrained features outperform other features, including ResNet-18 trained on ImageNet classification.}
\label{table:upmixing}
\end{table}

\subsection{Audio-Visual Source Separation} \label{sec:source-sep}
Next, we evaluate our pretrained audio-visual features on sound source separation, where the objective is to estimate individual sound sources based on their mixture using visual information. Since we are working with video datasets with spatial audio, the binaural cues could also be valuable in aiding separation; human listeners can distinguish sound sources not only based on differences in timbre, but also due to their spatial position. Therefore we hypothesize that a joint audio-visual representation that captures spatial information could improve performance on this task.

\paragraph{Model.} To test this hypothesis, we used the mix-and-separate framework \cite{Zhao18} adapted for stereo audio \cite{Gao19}. Our model takes mixed stereo audio tracks and visual frames as input and produces separated audio corresponding to the given visual stream (Figure \ref{fig:architectures}(d)), whereas Gao \etal \cite{Gao19} uses two visual streams for separation. \REMOVED{Similar to before, we}\ADDED{We} use a U-Net to generate a mask for the mixed audio input and produce the separated output. Pretrained \ADDED{audio-visual} features are \REMOVED{injected by concatenating them}\ADDED{concatenated} to the innermost layer of the network. 

\paragraph{Baselines and Evaluation Criteria.} We compare our pretrained audio-visual network \REMOVED{against}\ADDED{with} several baselines on the source separation task: (i) No audio-visual features; (ii) Our \REMOVED{audio-visual }network without any pretraining (``no pretraining"); \REMOVED{Audio-visual n}\ADDED{N}etworks trained on the \REMOVED{audio-visual }correspondence tasks of detecting (iii) mismatching semantic information (``mismatch task") \cite{Arandjelovi17} or (iv) shifted temporal alignment (``shift task") \cite{Korbar18,Owens18}, which use non-spatial audio cues to learn visual features; (v) ResNet-18 trained on ImageNet classification (``supervised"). For the training and evaluation criteria, we used the L1 distance between the output and target magnitude spectrograms \ADDED{averaged over time-frequency bins}. 

\begin{table}[!t]
\begin{center}
\begin{tabular}{|c|c|c|}
\hline
Visual sub-network & \asmr~ & FAIR-Play \\
\hline
No visual features & 0.0946 & 0.423 \\
No pretraining & 0.0953 & 0.422 \\
\hline
Mismatch task \cite{Arandjelovi17} & 0.0923 & 0.423 \\
Shift task \cite{Korbar18,Owens18} & 0.0918 & 0.422 \\
Flip task (ours) & 0.0898 & 0.410 \\
\hline
Supervised & 0.0885 & 0.362 \\
Ours + supervised & \textbf{0.0863} & \textbf{0.350} \\
\hline
\end{tabular}
\end{center}
\caption{Test set error of source separation performed on \asmr~ and FAIR-Play datasets. Our pretrained features \REMOVED{that capture spatial correspondence }outperform other self-supervised features and boost the performance of supervised ResNet-18 \REMOVED{visual }features.}
\label{table:source-sep}
\end{table}

\paragraph{Results.} The evaluation results on the test set are shown in Table \ref{table:source-sep}. The pretrained features from the audio-visual spatial correspondence task are useful for source separation on both the \asmr~ and the FAIR-Play datasets, outperforming all of the baselines except the strongly supervised ResNet-18 model trained on ImageNet. This observation is consistent with the fact that the source separation task depends on visually discriminating between different sound sources, for which ImageNet classification is well-suited. On the other hand, our features capture the spatial position of sound sources from both the audio and the visual frames, which may be helpful to the source separation task in complementary ways, \eg, if the mixed sounds are coming from two different locations. To determine if this is the case, we trained a new model by concatenating our audio-visual features with the pretrained ImageNet features (``Ours + supervised"). The addition of our features yielded a significant boost over using only the strongly supervised ResNet-18 features. \ADDED{The difference between this model and the rest is significant based on Wilcoxon signed-rank tests on \asmr~ ($p < 0.05$ for all tasks).} These results indicate that capturing audio-visual spatial correspondence is useful for sound source separation.


%% file: sections/reduced_ambisonic.tex
\section{Spatial Alignment in 360-Degree Video} \label{sec:ambisonics}

Our pretext task for learning spatial correspondence between sight and sound can be extended to 360-degree videos with full-sphere first-order ambisonics (FOA) audio. A key motivation for generalizing the task to this domain is learning a spatial representation of surround sound that can be applied to direction-of-arrival (DOA) estimation
\cite{adavanne2019multi}. 
However, \REMOVED{real sounds}\ADDED{recordings} with DOA annotations are extremely difficult to obtain, and currently training models relies on synthetic datasets for strongly supervised labels \cite{adavanne2018sound,adavanne2019multi, hirvonen2015classification,lopatka2016detection}. To address this constraint, we introduce a generalization of our audio-visual correspondence task that learns strong spatial audio cues from 360-degree videos with real spatial audio in a self-supervised manner. 

\paragraph{Problem Formulation.}
First-order ambisonics (FOA) extends stereo audio to the 3D setting, with extra channels to capture sound depth and height at time $t$: $a(t) = (a_w(t), a_y(t), a_z(t), a_x(t))$, where $a_w(t)$ represents omnidirectional sound pressure, and $(a_y(t), a_z(t), a_x(t))$ are front-back, up-down, and left-right sound pressure gradients respectively. FOA is often provided with 360-degree video to give viewers a full-sphere surround image and sound experience. Analogous to the case of stereo audio, we can train a model to detect whether the visual and audio streams are spatially aligned in 360-degree videos. To generate misaligned examples, we propose the transformation, 
$\tilde{a}(t) = (a_w(t),
   a_x(t)\sin{\theta} +  a_y(t)\cos{\theta}, 
   a_z(t),
   a_x(t)\cos{\theta} -  a_y(t)\sin{\theta})$
which rotates the audio about the z-axis by $\theta$. 

\paragraph{Implementation Details, Training and Results.} We use the same model architecture as for field-of-view video and stereo audio (depicted in Figure \ref{fig:architectures}). The main difference is that the input to the audio sub-network has the four FOA channels instead of two stereo channels. To train our model, we use the YouTube-360 dataset, which contains over a thousand 360-degree videos with FOA audio \cite{Morgado18}. Our model achieves about 60\% classification accuracy on the YouTube-360 test set; see Supplemental for details. 

\paragraph{Sound Localization.} Does the audio sub-network trained on the audio-visual spatial correspondence task learn a strong representation of 360-degree surround sound? We first investigate whether the audio embeddings extracted from held-out data contain 360-degree spatial audio cues. We observe that the audio features are strongly correlated with the directional energy of the sound based on an unsupervised projection using principal component analysis. Similar to Section \ref{sec:sound-source-loc}, we bin the embeddings based on their first two principal components and project the bin over the horizontal range of the video to track sound sources in the video using our self-supervised audio features\REMOVED{(Figure \ref{fig:360-loc-viz})}. We show qualitative results in Figure \ref{fig:asmr-audio-visual-loc}(c): notice that a moving vehicle is tracked from left to right.

To determine whether the learned spatial audio cues provide a quantitative boost on spatial audio localization tasks, we evaluated our audio sub-network on the Tau Spatial Sound dataset \cite{adavanne2019multi}. This dataset consists of 400 minute-long recordings of multiple sound events synthetically up-mixed to 36 different directions along the X-Y plane (every 10 degrees). \REMOVED{In the 2019 DCASE Sound Event Localization and Detection challenge \cite{adavanne2019multi}, participants were asked to output the class and direction-of-arrival (DOA) of these sound events. All of the models were trained from scratch in a strongly supervised manner.} \ADDED{Our objective is to predict the direction-of-arrival (DOA) of sound events given FOA audio input.} \REMOVED{We conducted two experiments on the Tau Spatial Sound dataset.} \ADDED{We first}\REMOVED{First, we} compared a baseline model trained from scratch with one that was initialized using the weights from our pretrained audio sub-network. We found that our pretrained weights provided a notable boost over the baseline, using the error in the azimuth angle as the evaluation criteria (Table \ref{table:dcase-doa}). \REMOVED{Second}\ADDED{Next}, we extracted the embeddings of the Tau Spatial Sound audio clips using our pretrained audio sub-network and performed classification using a linear SVM, providing only one example of an event from each azimuth angle (one-shot learning). Even in this case, we were able to predict the direction of arrival with a mean error of only about 30 degrees. These results suggest that the spatial information extracted using our self-supervised task are useful for acoustic localization.

\begin{table}[t]
\begin{center}
\begin{tabular}{|c|c|}
\hline
Method & Mean Error \\
\hline
Trained from scratch & 20.5 \\ 
Pretrained weights & \textbf{17.5} \\ 
\hline
One-shot SVM on random embeddings & 78.6 \\
One-shot SVM on our embeddings & \textbf{29.5} \\
\hline
\end{tabular}
\end{center}
\caption{\REMOVED{Performance on DCASE DOA estimation}\ADDED{DOA estimation performance on the Tau Spatial Sound dataset}. Error shown is mean prediction error in degrees (lower is better, maximum is 180, random is 90). Standard deviation of the top models are 1.17 and 0.62 based on 4-fold cross validation. }
\label{table:dcase-doa}
\end{table}

%% file: sections/conclusion.tex
\section{Conclusion}

We have demonstrated a simple, yet effective, approach for self-supervised representation learning from video with spatial audio and its application to three downstream audio-visual tasks. Critical to our approach was the ability to train on a large corpus of video with spatial audio. Our work opens up the possibility of exploring network architectures and cross-modal self-supervised training losses~\cite{Tian2019ContrastiveMC} that jointly leverage the spatial and semantic cues present in the visual and spatial audio channels in an effective manner.